\documentclass[conference]{IEEEtran}
\usepackage[T1]{fontenc}
\usepackage{amsmath,amssymb}
\usepackage{graphicx}
\usepackage{booktabs}
\usepackage{siunitx}
\usepackage{microtype}
\usepackage{xurl}
\usepackage[hidelinks]{hyperref}

\usepackage[capitalise,noabbrev]{cleveref}

\title{Is Token Confidence Just Chain Length?\\ A Signal-Processing Audit of Reasoning Traces}
\author{\IEEEauthorblockN{Pragna Mallappa}
\IEEEauthorblockA{Keshav Memorial Institute of Technology\\ mallappa.pragna2197@gmail.com}}

\begin{document}
\maketitle

\begin{abstract}
Reasoning language models answer math problems by writing long chains of thought, and sampling many chains per problem and taking a majority vote raises accuracy at a large cost in generated text. A popular way to cut that cost scores each chain by the model's confidence in each token it writes, where a token is a word or part of a word. The method averages this confidence over a sliding window of 2048 tokens and keeps or stops chains based on the lowest average. We treat the confidence, recorded token by token, as a signal in the sense of audio or radio, and examine the method with standard signal-processing tools such as spectral analysis and statistical tests for detecting a sudden change. We sampled 16 chains for each of 500 problems from two math benchmarks with a small open reasoning model. The windowed-minimum score turns out to measure mostly how long a chain is. Ranking chains by length alone does as well as or better than the confidence score, and once length is accounted for the score carries no information about correctness. Signal processing explains why: the minimum of a longer stretch of a slowly drifting signal is lower by chance alone, and the confidence signal drifts slowly enough that this effect dominates. The spectrum does hold a small signal that length does not explain, since incorrect chains carry more rapid token-to-token fluctuation, but it is too weak to improve the vote. For stopping chains early, neither the sliding average nor a classical change detector beats the simpler choice of sampling fewer chains.
\end{abstract}

\section{Introduction}

Large language models now solve many math problems by first writing out a long chain of reasoning, called a chain of thought, and only then stating an answer~\cite{wei2022cot,Guo_2025}. Because the model samples each token at random from its predicted distribution, two chains for the same problem can reach different answers. Their failures are not spread evenly either: on grade-school word problems, a model can be nearly perfect at step-by-step arithmetic yet fail systematically on problems that require counting under restrictions~\cite{parupudi2025brittle}. A simple and strong way to exploit this is to sample several chains and return the most common answer, a method known as self-consistency or majority voting~\cite{wang2023selfconsistency}. Its cost is that each extra chain adds thousands of generated tokens. This has motivated methods that try to tell good chains from bad ones, so that bad chains can be left out of the vote or stopped before they finish~\cite{aggarwal2023adaptive,li2024esc}.

One family of such methods uses nothing but the model's own confidence~\cite{kadavath2022know,fu2025deepconf}. At every step the model assigns a probability to each possible next token. When this probability is concentrated on one candidate the model is sure about what comes next, and when it is spread over several candidates it is not. The shape of this distribution says more than its top candidate alone, and how it changes under different sampling temperatures has been used to judge generated text without any reference answer~\cite{parupudi2026temperature}. DeepConf~\cite{fu2025deepconf} turns this into one number per token, called the token confidence. It scores a chain by averaging token confidence over a sliding window of 2048 tokens and keeping the lowest average. The score is used in two ways. Offline, low-scoring chains are dropped before the vote. Online, a chain is stopped as soon as its windowed confidence falls below a threshold that is set from a few warmup chains.

This paper looks at the same computation through the lens of signal processing. The token confidences of one chain form a sequence indexed by position, which is a discrete-time signal in the same sense as a sampled audio waveform. Averaging over a sliding window is a moving-average filter, often called a boxcar filter because all of its weights are equal. It is a low-pass filter, meaning that it keeps slow changes in the signal and suppresses fast ones. Keeping the lowest filtered value, or stopping when the filtered value crosses a threshold, is a detector. Seen this way, a boxcar of length 2048 is a single point in a large space of filters and detectors, and nothing about the confidence signal was measured to arrive at it. Signal processing offers standard tools for measuring such a signal and for designing filters and detectors that match what is measured.

We carry out this measurement on 8000 chains from a small open reasoning model on two math benchmarks, and the answer is mostly negative. The confidence signal has a power spectrum close to $1/f$, which means that most of its variance sits in slow drifts. A long moving average keeps those drifts, so the minimum of the filtered signal keeps falling as a chain gets longer, even within a single chain. As a result DeepConf's score is largely a measure of chain length, and length is known to predict correctness on its own~\cite{hassid2025shorter}. Ranking chains by length matches or beats every confidence score we test, and once length is removed the lowest group confidence is at chance on both benchmarks. This fits a broader pattern in the evaluation of language models, where an apparent difference turns out to come from how something is measured, as when gaps between models on leaderboards come mostly from how answers are scored~\cite{parupudi2026harness}.

Our contributions are the following. We recast confidence-based chain selection as filtering followed by detection, which makes its design choices explicit and testable (\cref{sec:signals}). We measure the spectrum of the confidence signal and find a small effect that length does not explain: incorrect chains carry relatively more power at the fastest fluctuations, and correct chains more at periods of about 18 tokens (\cref{sec:spectrum}). We show, with surrogate signals that keep or destroy the spectrum, that the length dependence of the lowest group confidence follows from the spectrum and the minimum operation (\cref{sec:selection}). Finally, we cast early stopping as change detection and find that neither the boxcar threshold nor a CUSUM detector beats sampling fewer chains at equal cost (\cref{sec:online}).

\section{Reasoning traces as signals}
\label{sec:signals}

\subsection{The confidence signal}
Consider a chain of $T$ generated tokens. At position $t$ the model outputs a probability distribution over its vocabulary. Following DeepConf, we take the $k$ most likely candidates, with probabilities $p_t^{(1)} \ge \dots \ge p_t^{(k)}$, and define the token confidence as
\begin{equation}
  c_t = -\frac{1}{k}\sum_{j=1}^{k} \log p_t^{(j)} .
  \label{eq:conf}
\end{equation}
When the distribution is peaked, nearly all of the probability sits on the top candidate, which leaves tiny probabilities, and so very negative logarithms, for the other candidates. Token confidence is therefore high when the model is sure of the next token and low when several candidates compete. We use $k=20$. The sequence $c_1,\dots,c_T$ is the \emph{confidence signal} of the chain, and the rest of the paper treats it as a discrete-time signal.

\subsection{DeepConf as a filter and a detector}
DeepConf's group confidence at position $t$ is the mean of the last $w$ token confidences. In filter terms this is $y = h * c$, where $*$ denotes convolution and $h$ is a boxcar of length $w$ whose weights all equal $1/w$. The score of the chain, which DeepConf calls the lowest group confidence, is $s = \min_t y_t$.

\begin{figure}[t]
  \centering
  \includegraphics[width=\columnwidth]{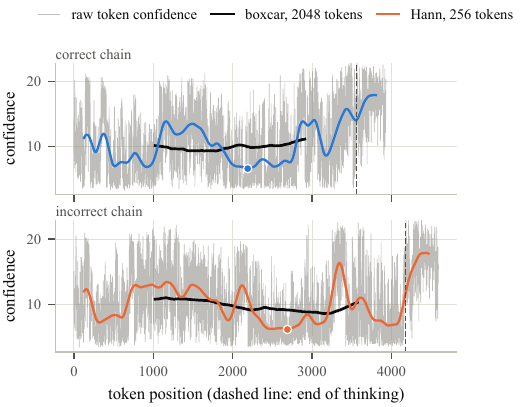}
  \caption{Token confidence of one correct and one incorrect chain for the same MATH500 problem. The raw signal (gray) fluctuates strongly from token to token. The 2048-token boxcar used by DeepConf (black) keeps only the slowest drift, while a 256-token Hann window (color) follows dips of a few hundred tokens. Dots mark the minimum of the Hann output.}
  \label{fig:signal}
\end{figure}

A filter can be described by its frequency response, which says how strongly it passes a slow or a fast oscillation. Frequency has a concrete meaning here. A component of the confidence signal at $f$ cycles per token rises and falls once every $1/f$ tokens, so low frequencies are slow drifts over hundreds of tokens and high frequencies are jitter from one token to the next. The boxcar's response first reaches zero at $f = 1/w$, so with $w = 2048$ it passes only changes that unfold over thousands of tokens. A chain shorter than the window yields a single window, and its score is then simply its mean confidence. This case is common in our data: 44\% of the answered MATH500 chains and 75\% of the answered GSM8K chains are shorter than 2048 tokens. \Cref{fig:signal} shows the confidence signal of two chains together with the boxcar output and a shorter, smoother alternative.

\subsection{Alternatives from signal processing}
\label{sec:alternatives}
\emph{Window shape and length.} The Hann and Gaussian windows taper their weights toward the edges. This gives them much smaller sidelobes than the boxcar, which means that less fast fluctuation leaks into the smoothed output. We also test the exponential moving average $y_t = (1-\alpha)\,y_{t-1} + \alpha\,c_t$ with $\alpha = 1/\tau$. It is a recursive low-pass filter that weights recent tokens most and needs no buffer of past values. Each filter is followed by one of two pooling rules: the minimum, as in DeepConf, or the mean of the lowest 10\% of filtered values.

\emph{Spectral features.} We estimate the power spectral density (PSD), which describes how the variance of a signal is divided among frequencies, with Welch's method~\cite{welch1967psd}. It splits the signal into overlapping segments of 256 tokens, computes the squared magnitude of each segment's Fourier transform, and averages the results. Summing the PSD within eight bands spaced evenly on a log scale from $1/256$ to $1/2$ cycles per token gives eight band powers. A weighted sum of log band powers is a filter-bank energy detector: a set of band-pass filters, each followed by an energy measurement, whose outputs are combined linearly. We learn the weights with logistic regression and evaluate it with five-fold cross-validation across problems, so that no problem is used for both training and testing. Before fitting, each feature is standardized within each problem, so that the detector compares chains of the same problem rather than problems with each other. We also test a fixed version that needs no training, the high-frequency energy detector, which scores a chain by the negative log power in the top band, from about 0.27 to 0.5 cycles per token.

\emph{Change detection.} For online stopping we use Page's cumulative sum test, known as CUSUM~\cite{page1954cusum,basseville1993detection}. A few warmup chains give the baseline mean $\mu_0$ and standard deviation $\sigma_0$ of token confidence for the problem, and each new chain is standardized as $z_t = (c_t - \mu_0)/\sigma_0$. The detector keeps the statistic
\begin{equation}
  S_t = \max\left(0,\; S_{t-1} - z_t - \kappa\right), \qquad S_0 = 0,
  \label{eq:cusum}
\end{equation}
and stops the chain when $S_t$ exceeds a level $h$. The statistic grows while confidence stays below the baseline and returns to zero when confidence recovers, so it accumulates evidence of a sustained drop and forgets isolated dips. The drift $\kappa$ sets the smallest drop worth detecting and $h$ sets how much evidence is needed before an alarm. CUSUM is the classical detector for an abrupt drop in the mean of a noisy signal~\cite{basseville1993detection}, which is the event that a confidence threshold is meant to catch.

\section{Experimental setup}
\label{sec:setup}
We use DeepSeek-R1-Distill-Qwen-1.5B~\cite{Guo_2025}, a reasoning model with 1.5 billion parameters that was trained to imitate the chains of thought of DeepSeek-R1. We take the first 200 problems of MATH500~\cite{lightman2023verify,hendrycks2021math}, a set of competition math problems, and the first 300 problems of the GSM8K test set~\cite{cobbe2021gsm8k}, a set of grade-school word problems. For each problem we sample 16 chains at temperature 0.6 with nucleus sampling at $p=0.95$, allow at most 8192 generated tokens, and ask for the final answer in a \texttt{\textbackslash boxed\{\}} command. We read the answer from the text after the model's end-of-thinking marker and check it against the reference with the \texttt{math-verify} library. A chain that reaches the token limit before finishing its reasoning has no answer and counts as wrong. Generation used vLLM~\cite{kwon2023vllm} on a single RTX 4090 GPU and logged the top 20 log-probabilities at every position. The two runs produced 22.0 million tokens in 84 minutes of generation. \Cref{tab:data} summarizes them.

\begin{table}[t]
  \centering
  \caption{The two runs. Accuracy is over all chains, with chains that hit the token limit counted as wrong. Mixed problems have at least one correct and one incorrect answered chain, and only they enter the AUROC.}
  \label{tab:data}
  \begin{tabular}{@{}lrr@{}}
    \toprule
    & MATH500 & GSM8K \\
    \midrule
    Problems $\times$ chains & $200 \times 16$ & $300 \times 16$ \\
    Mean tokens per chain & 3761 & 2081 \\
    Chains hitting the token limit & 17.7\% & 4.0\% \\
    Accuracy of one chain & 78.6\% & 86.2\% \\
    Majority vote of 16 chains & 92.0\% & 91.0\% \\
    Any of 16 chains correct & 92.5\% & 98.0\% \\
    Mixed problems & 31 & 96 \\
    \bottomrule
  \end{tabular}
\end{table}

\emph{Metrics.} Our main measure of how well a score separates good chains from bad ones is the within-problem AUROC. For a problem whose answered chains include both correct and incorrect ones, it is the probability that a randomly chosen correct chain receives a higher score than a randomly chosen incorrect chain. We average it over all such problems, and we give 95\% confidence intervals from a bootstrap that resamples problems. Comparing chains within a problem removes the effect of problem difficulty, and restricting to answered chains removes the trivial signal of hitting the token limit. To ask whether a score knows anything beyond chain length, we also compute a length-removed AUROC. Within each problem we subtract the mean from both the score and the log of the chain length, fit a single slope that predicts the first from the second across all problems, and compute the AUROC of what the slope does not explain. We also report the accuracy of filtering before voting: keep the highest-scoring fraction $\eta$ of a problem's answered chains and take the majority answer among them. For online stopping we report accuracy against the fraction of tokens spent, relative to generating all 16 chains in full.

\emph{Baselines.} Shorter chains are known to be correct more often~\cite{hassid2025shorter,chen2024overthinking}, so chain length is a baseline for every result. We also compare with the mean confidence of the chain and with DeepConf's three offline scores: the lowest group confidence, the mean of the lowest 10\% of group confidences, and the mean confidence of the last 2048 tokens. For online stopping we compare with two budget rules that ignore confidence entirely: sampling fewer chains, and cutting every chain at a fixed token cap~\cite{muennighoff2025s1}.

\section{What the spectrum shows}
\label{sec:spectrum}

\begin{figure*}[t]
  \centering
  \includegraphics[width=\textwidth]{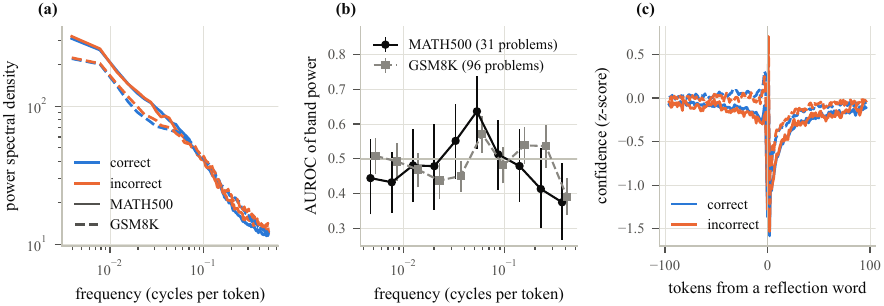}
  \caption{Spectrum of the confidence signal over the thinking part of answered chains, restricted to mixed problems. (a)~Mean power spectral density of correct and incorrect chains. Both fall roughly as $f^{-0.75}$, so most of the variance sits in slow drifts. (b)~Within-problem AUROC of the power in each of ten log-spaced bands, where values above 0.5 mean that more power goes with correct chains. Bars are 95\% bootstrap intervals over problems. On both benchmarks, power near 0.06 cycles per token goes with correct chains and power in the fastest band with incorrect ones. (c)~Average standardized confidence around a reflection word such as ``Wait.'' Confidence drops sharply at the word and recovers over about 50 tokens, with the same shape for correct and incorrect chains.}
  \label{fig:spectrum}
\end{figure*}

For this section we use only the thinking part of each chain, the text before the end-of-thinking marker, so that the short final answer does not mix into the spectrum. \Cref{fig:spectrum}a shows that the confidence signal is far from white noise, a signal whose power is spread evenly over all frequencies. Its PSD falls steadily with frequency, and a straight-line fit on log axes gives a slope of $-0.80$ for correct and $-0.76$ for incorrect MATH500 chains, and $-0.70$ and $-0.68$ on GSM8K. A slope near $-1$ is the signature of a $1/f$ process, whose slow components are strong and whose values stay correlated over long stretches. This matters for the next section, because a low-pass filter removes little of the variance of such a signal.

The spectra of correct and incorrect chains nearly coincide, but they differ in a small and consistent way. The slope of incorrect chains is less steep, by 0.041 (95\% interval 0.018 to 0.064) on MATH500 and by 0.020 (0.007 to 0.032) on GSM8K. In other words, incorrect chains carry relatively more of their variance in fast fluctuations. \Cref{fig:spectrum}b locates the difference. On both benchmarks the band around 0.06 cycles per token, which corresponds to a period of about 18 tokens, has more power in correct chains, with an AUROC of 0.64 (0.53 to 0.74) on MATH500 and 0.57 (0.52 to 0.62) on GSM8K. The fastest band, with periods of two to three tokens, has more power in incorrect chains, with an AUROC of 0.38 (0.27 to 0.49) and 0.39 (0.34 to 0.44). A period of 18 tokens is roughly the length of a short sentence or an equation step, so one reading is that correct chains move through their work with a steadier step-by-step rhythm. We have not tested this reading.

Reasoning models often write words like ``Wait'' or ``Alternatively'' when they revisit their own work. We call these reflection words and average the standardized confidence signal in a window around each occurrence, which estimates the typical response of the signal to such an event. \Cref{fig:spectrum}c shows that confidence falls sharply at the reflection word, to about 1.3 standard deviations below the chain's mean on MATH500 and 1.0 on GSM8K, and recovers over about 50 tokens. The response has the same shape in correct and incorrect chains. What differs is how often it occurs: incorrect chains contain more reflection words per thousand tokens, 6.7 against 5.8 on MATH500 (difference 0.9, interval 0.0 to 1.8) and 7.7 against 6.9 on GSM8K (difference 0.8, interval 0.4 to 1.3). Each reflection word leaves a dip lasting tens of tokens, so a chain that hesitates more often also has more and deeper dips in its smoothed confidence.

\section{Filters for choosing chains}
\label{sec:selection}

\begin{figure*}[t]
  \centering
  \includegraphics[width=\textwidth]{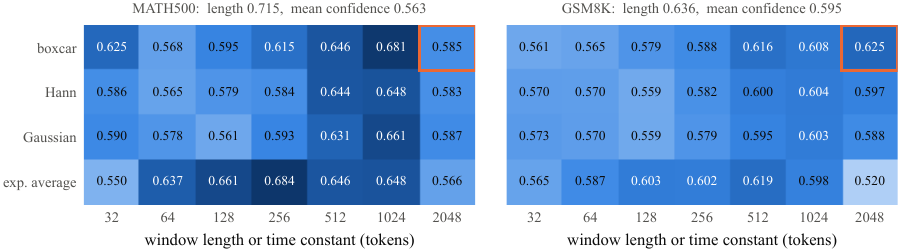}
  \caption{Within-problem AUROC of the minimum of the filtered confidence signal, for four filter families and seven window lengths. The orange box marks DeepConf's 2048-token boxcar. The titles give the AUROC of two simple baselines: ranking chains from shortest to longest, and their mean confidence. Shorter windows help on MATH500 but not on GSM8K, and no filter beats chain length.}
  \label{fig:filters}
\end{figure*}

From here on, scores are computed over the whole chain, as in DeepConf. \Cref{fig:filters} compares filter families and window lengths, each followed by the minimum. On MATH500, windows shorter than DeepConf's do better. The exponential average with $\tau = 256$ reaches an AUROC of 0.684 against 0.585 for the 2048-token boxcar, a gain of 0.100 (0.003 to 0.213). On GSM8K the pattern disappears, and no filter beats DeepConf's own setting of 0.625. The window shape matters less than its length, and the tapered Hann and Gaussian windows are usually slightly worse than a boxcar of the same length. The learned filter-bank detector reaches 0.566 on MATH500 and 0.644 on GSM8K, and neither differs reliably from DeepConf.

\begin{figure*}[t]
  \centering
  \includegraphics[width=\textwidth]{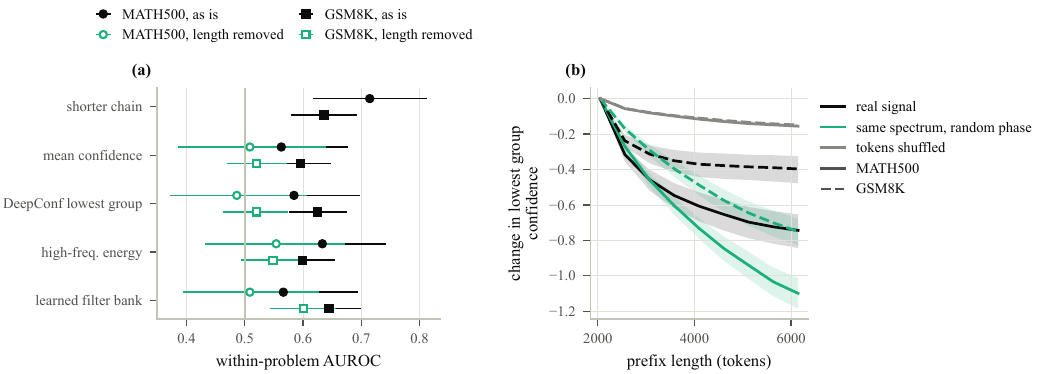}
  \caption{How much of each score is chain length. (a)~Within-problem AUROC with 95\% bootstrap intervals, before (black) and after (green) removing the part of the score that the log of chain length explains. DeepConf's lowest group confidence falls to chance, while the two spectral detectors keep part of their signal. (b)~Within answered chains of at least 6144 tokens, the change in lowest group confidence as the prefix grows from 2048 tokens. The real signal falls far more than a token-shuffled copy with the same values, and about as much as or less than a copy with the same spectrum and random phases. Shading shows 95\% bootstrap intervals over chains.}
  \label{fig:length}
\end{figure*}

The more telling comparison is with the baseline in the panel titles. Ranking chains from shortest to longest gives an AUROC of 0.715 on MATH500, which beats DeepConf by 0.130 (0.019 to 0.257), and 0.636 on GSM8K, which ties it. \Cref{fig:length}a asks how much of each score is length. Within a problem, the lowest group confidence has a rank correlation with chain length of $-0.53$ on MATH500 and $-0.75$ on GSM8K. Once length is removed its AUROC falls to 0.486 (0.372 to 0.607) and 0.520 (0.463 to 0.575), which is chance on both benchmarks. Mean confidence behaves the same way. The spectral detectors keep part of their signal. The high-frequency energy detector, which has no fitted weights, keeps an AUROC of 0.554 on MATH500 and 0.549 (0.494 to 0.599) on GSM8K. The learned filter bank keeps 0.601 (0.543 to 0.657) on GSM8K but only 0.509 on MATH500, where it was trained on about 25 problems per fold. Adding the high-frequency score to the length score, with equal weight after standardizing both within each problem, does not raise the AUROC above length alone on either benchmark.

Signal processing explains where the length dependence comes from. The minimum of a noisy signal over a longer stretch is lower, because a longer stretch gives the signal more chances to wander down. How strong this effect is after filtering depends on the spectrum. Averaging $w$ samples of white noise shrinks its standard deviation by a factor of $\sqrt{w}$, about 45 for $w = 2048$, so the minimum of the filtered noise barely moves as the stretch grows. A $1/f$ signal keeps most of its slow drift through the same filter, so its filtered minimum keeps falling. \Cref{fig:length}b tests this within single chains, which removes every difference between chains. As the prefix of a chain grows from 2048 to 6144 tokens, its lowest group confidence falls by 0.74 (0.65 to 0.84) on MATH500 and by 0.40 (0.33 to 0.48) on GSM8K. Shuffling the tokens of the same chain, which keeps the values but makes the spectrum flat, shrinks the fall to 0.15. Randomizing the Fourier phases instead, which keeps the spectrum exactly, gives a fall of 1.10 and 0.75. The spectrum alone therefore predicts a fall at least as large as the one observed. The same effect shows up across chains with the same outcome. Among correct MATH500 chains, the rank correlation with length is $-0.50$ for the lowest group confidence but only $-0.08$ for mean confidence, so the minimum, not a lower confidence level, creates most of the dependence. On GSM8K mean confidence also falls with length ($-0.50$), and most chains are shorter than one window, so there the two scores largely coincide.

None of this changes the final answer much. Keeping the top 6\% to 75\% of answered chains by any score and voting among them gives accuracies between 89.0\% and 92.0\% on MATH500 and between 87.7\% and 93.3\% on GSM8K, against 92.0\% and 91.0\% for the full vote. No score gains consistently across keep fractions or benchmarks. Selection can only matter on mixed problems, and on most of those the majority of chains is already correct.

\section{Stopping chains early}
\label{sec:online}

\begin{figure*}[t]
  \centering
  \includegraphics[width=\textwidth]{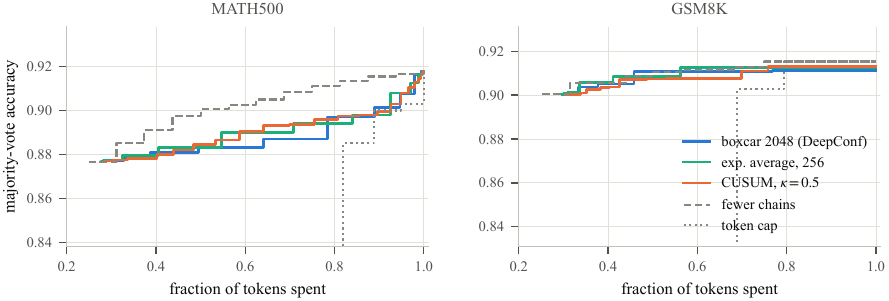}
  \caption{Majority-vote accuracy against the fraction of tokens spent, relative to running all 16 chains to the end, for early-stopping rules simulated on the generated chains. Each curve keeps the best accuracy reached at or below each cost. Simply sampling fewer chains (dashed) matches or beats every detector, including the 2048-token boxcar and CUSUM.}
  \label{fig:online}
\end{figure*}

We simulate early stopping on the finished chains. For each problem, four warmup chains run to the end and set the baseline mean and standard deviation of token confidence. Each of the other 12 chains is standardized with this baseline and watched by a detector, which may stop it after its first 128 tokens. A stopped chain costs the tokens it has produced and gets no vote. The answer is the majority among the surviving answered chains and the warmup chains. We average over ten random choices of warmup chains, and we trace out the trade-off between cost and accuracy by sweeping each detector's threshold. The level detectors stop a chain when a boxcar or exponential average of the standardized signal falls below a threshold, which includes DeepConf's online rule with its 2048-token window as one case. The CUSUM detector of \cref{eq:cusum} stops it when $S_t$ exceeds $h$.

\Cref{fig:online} shows the result. On MATH500, sampling fewer chains gives the best accuracy at every cost. At half the full cost it reaches 90.0\%, against 88.4\% for the best detector, and at 70\% of the cost it reaches 90.8\%, against 89.4\% for CUSUM and 88.7\% for the 2048-token boxcar. On GSM8K all methods lie within 0.6 points of each other at every cost from 40\% to 70\%, which is about two problems in 300. CUSUM, the detector designed for a sustained drop in the mean, does no better than a plain threshold on a moving average. A fixed token cap is far worse on both benchmarks, because it removes long chains that would have voted correctly. These differences are small in absolute terms, and with 200 and 300 problems we cannot resolve gaps below about one point. The conclusion we draw is limited to this: confidence-based stopping offers no advantage over the simplest way to spend fewer tokens.

\section{Related work}

Self-consistency~\cite{wang2023selfconsistency} samples several chains of thought~\cite{wei2022cot} and takes the majority answer. Adaptive variants stop sampling once the votes so far agree~\cite{aggarwal2023adaptive,li2024esc}, which saves chains rather than tokens within a chain. Using the model's own probabilities to judge its answers goes back at least to work on calibration of language models~\cite{kadavath2022know}, and semantic entropy~\cite{kuhn2023semantic} measures uncertainty across sampled answers rather than within one chain. DeepConf~\cite{fu2025deepconf} is the method we examine. It filters chains by their windowed token confidence and reports gains for large models on competition math. A separate line of work finds that reasoning models often overthink~\cite{chen2024overthinking} and that their shorter chains are more often correct~\cite{hassid2025shorter}, and budget forcing~\cite{muennighoff2025s1} controls the length of the thinking phase directly. Our results connect these two lines: on the chains we study, much of what confidence filtering detects is the length effect.

Other work looks inside reasoning instead of scoring only the final answer. Parupudi~\cite{parupudi2025brittle} sorts the sentences of GSM8K reasoning chains into reasoning modes and finds that accuracy collapses on some modes while staying near perfect on others. That analysis reads the content of the text, while ours ignores the text and uses only the model's token probabilities, so the two views are complementary. Features of the next-token distribution have also been studied outside reasoning. Parupudi et al.~\cite{parupudi2026temperature} show that how temperature reshapes this distribution predicts the creativity of generated text better than standard measures such as perplexity and entropy. They also find that per-token measures cannot separate nearby temperatures and point to sequence-level features instead. Our token confidence is a per-token measure of this kind, and its spectrum is one sequence-level summary of it. Finally, measurement choices can create apparent effects. Parupudi~\cite{parupudi2026harness} shows that the ranking of language models on multiple-choice leaderboards changes with the evaluation setup, and that a small set of items whose outcome depends on that setup produces most of the gaps between models. The length effect we report is a similar case: a property of how chains are scored that can pass for a signal about correctness.

On the signal-processing side we use textbook tools, namely Welch's spectral estimate~\cite{welch1967psd} and Page's CUSUM test~\cite{page1954cusum,basseville1993detection}.

\section{Limitations and conclusion}

Our study uses one small model, a distilled 1.5-billion-parameter model, and two math benchmarks. DeepConf reports its gains for much larger models and harder problems, where chains are longer and the relation between length and correctness may differ, so our results do not show that confidence filtering fails in that setting. They show that on our data its score does not add to length. The number of mixed problems is modest, 31 on MATH500 and 96 on GSM8K, which leaves wide intervals on MATH500. We used one sampling temperature and one value of $k$ in the confidence definition. Temperature reshapes the next-token distribution~\cite{parupudi2026temperature}, so the confidence signal and its spectrum may well change with it. Our answers are numbers that can be checked exactly, and extending the analysis to free-form answers would require grouping them by meaning before voting~\cite{kuhn2023semantic}. Finally, our early-stopping results are simulated on finished chains. A real system would stop generation, which saves time as well as tokens but does not change the votes.

Treating token confidence as a signal turns a heuristic into a filter followed by a detector, and that makes it possible to ask what each part does. On our data the answer is that the 2048-token boxcar and the minimum together measure chain length, for a reason that follows from the slowly drifting spectrum of the signal. The spectrum also holds a weak signal of its own, in the balance between fast jitter and step-sized fluctuations, which length does not explain. We suggest that work on confidence-based chain selection report chain length as a baseline and report AUROC with length removed, since without both a length effect can pass for a confidence effect.

\bibliographystyle{IEEEtran}
\bibliography{refs}
\end{document}